\documentclass[10pt,journal,compsoc]{IEEEtran}
\renewcommand{\baselinestretch}{0.98}
\ifCLASSOPTIONcompsoc
  \usepackage[nocompress]{cite}
\else
  \usepackage{cite}
\fi

\usepackage{graphicx}
\usepackage{comment}
\usepackage{url}
\usepackage{amsmath,amssymb} % define this before the line numbering.
\usepackage{color}

\usepackage{multirow}
\usepackage{multicol}
\usepackage{floatrow}
\usepackage[noend]{algpseudocode}
\usepackage{algorithmicx,algorithm}
\usepackage{bbding}

\usepackage{tikz,xcolor,hyperref}% Make Orcid icon
\usepackage{subfigure}
\usepackage{booktabs}
\usepackage{makecell}
\usepackage{xspace}

\definecolor{lime}{HTML}{A6CE39}
\DeclareRobustCommand{\orcidicon}{%
    \begin{tikzpicture}
    \draw[lime, fill=lime] (0,0) 
    circle [radius=0.16] 
    node[white] {{\fontfamily{qag}\selectfont \tiny ID}};    \draw[white, fill=white] (-0.0625,0.095) 
    circle [radius=0.007];    \end{tikzpicture}
    \hspace{-2mm}}
\foreach \x in {A, ..., Z}{%
    \expandafter\xdef\csname orcid\x\endcsname{\noexpand\href{https://orcid.org/\csname orcidauthor\x\endcsname}{\noexpand\orcidicon}}
    }

\begin{document}
\title{Path to Medical AGI: Unify Domain-specific Medical LLMs with the Lowest Cost}

% \author{Michael~Shell,~\IEEEmembership{Member,~IEEE,}
%         John~Doe,~\IEEEmembership{Fellow,~OSA,}
%         and~Jane~Doe,~\IEEEmembership{Life~Fellow,~IEEE}% <-this % stops a space
% \author{Chong~Mou, Jian~Zhang
\author{Juexiao~Zhou$^{1,2,\#}$, Xiuying Chen$^{1,2,\#}$, Xin~Gao$^{1,2,*}$
\thanks{
$^1$Computer Science Program, Computer, Electrical and Mathematical Sciences and Engineering Division, King Abdullah University of Science and Technology (KAUST), Thuwal 23955-6900, Kingdom of Saudi Arabia\\
$^2$Computational Bioscience Research Center, King Abdullah University of Science and Technology (KAUST), Thuwal 23955-6900, Kingdom of Saudi Arabia\\
$^\#$These authors contributed equally.\\
$^*$Corresponding author. e-mail: xin.gao@kaust.edu.sa\\
}}


\markboth{}%
\\
\IEEEtitleabstractindextext{%
\begin{abstract}
Medical artificial general intelligence (AGI) is an emerging field that aims to develop systems specifically designed for medical applications that possess the ability to understand, learn, and apply knowledge across a wide range of tasks and domains. 
Large language models (LLMs) represent a significant step towards AGI. 
However, training cross-domain LLMs in the medical field poses significant challenges primarily attributed to the requirement of collecting data from diverse domains. 
This task becomes particularly difficult due to privacy restrictions and the scarcity of publicly available medical datasets.
Here, we propose Medical AGI (MedAGI), a paradigm to unify domain-specific medical LLMs with the lowest cost, and suggest a possible path to achieve medical AGI. With an increasing number of domain-specific professional multimodal LLMs in the medical field being developed, MedAGI is designed to automatically select appropriate medical models by analyzing users' questions with our novel adaptive expert selection algorithm. 
It offers a unified approach to existing LLMs in the medical field, eliminating the need for retraining regardless of the introduction of new models. 
This characteristic renders it a future-proof solution in the dynamically advancing medical domain.
To showcase the resilience of MedAGI, we conducted an evaluation across three distinct medical domains: dermatology diagnosis, X-ray diagnosis, and analysis of pathology pictures. 
The results demonstrated that MedAGI exhibited remarkable versatility and scalability, delivering exceptional performance across diverse domains.
Our code is publicly available to facilitate further research at \url{https://github.com/JoshuaChou2018/MedAGI}.

\end{abstract}

\begin{IEEEkeywords}
Healthcare, Deep learning, Large language model, Artificial general intelligence
\end{IEEEkeywords}}

\maketitle

\IEEEdisplaynontitleabstractindextext

% \fi
\IEEEpeerreviewmaketitle

\section{Introduction}
Artificial General Intelligence (AGI)\cite{goertzel2014artificial} refers to highly autonomous systems that possess the ability to understand, learn, and apply knowledge across a wide range of tasks and domains.
These systems are designed to match or even exceed human competencies in intellectual tasks. 
Essentially, AGI represents the pinnacle objective within the field of artificial intelligence (AI)\cite{winston1984artificial}.
Within this realm, Medical AGI is an emerging field that aims to develop Artificial General Intelligence systems specifically designed for medical applications, encompassing tasks such as disease diagnosis, treatment planning, and patient care optimization. 
Large language models (LLMs)\cite{bubeck2023sparks} represent a significant step towards AGI by showcasing the power of language processing and understanding. 
During the past few months, significant progress has been made in the field of LLMs, revolutionizing language comprehension and enabling complex linguistic tasks\cite{kung2023performance, sallam2023chatgpt}. 
Among the highly anticipated models, ChatGPT, developed by OpenAI, has garnered attention for its exceptional capabilities. 
This model is especially proficient in generating human-like text based on the input it receives, demonstrating an impressive understanding of nuanced contexts and varied linguistic styles.
Specifically, ChatGPT shows great potential in Medical AGI by assisting with medical disease diagnosis through patients conversations, such as ophthalmic diagnosis\cite{balas2023conversational},  pathology diagnosis\cite{sinha2023applicability}, and health care discussion\cite{vaishya2023chatgpt}.
% mijwil2023chatgpt ufuk2023role hu2023advancing beltrami2023consulting
% GPT-4\cite{bubeck2023sparks} has garnered attention for its exceptional capabilities, such as generating detailed image descriptions, explaining unusual visual occurrences, creating websites based on textual descriptions, and even functioning as a virtual family doctor\cite{lee2023benefits}. 
% However, some features of GPT-4 remain inaccessible to the public, as they are closed-source and require an application to access through the API. 

\begin{figure*}[!htb]
    \centering
    \includegraphics[width=1\linewidth]{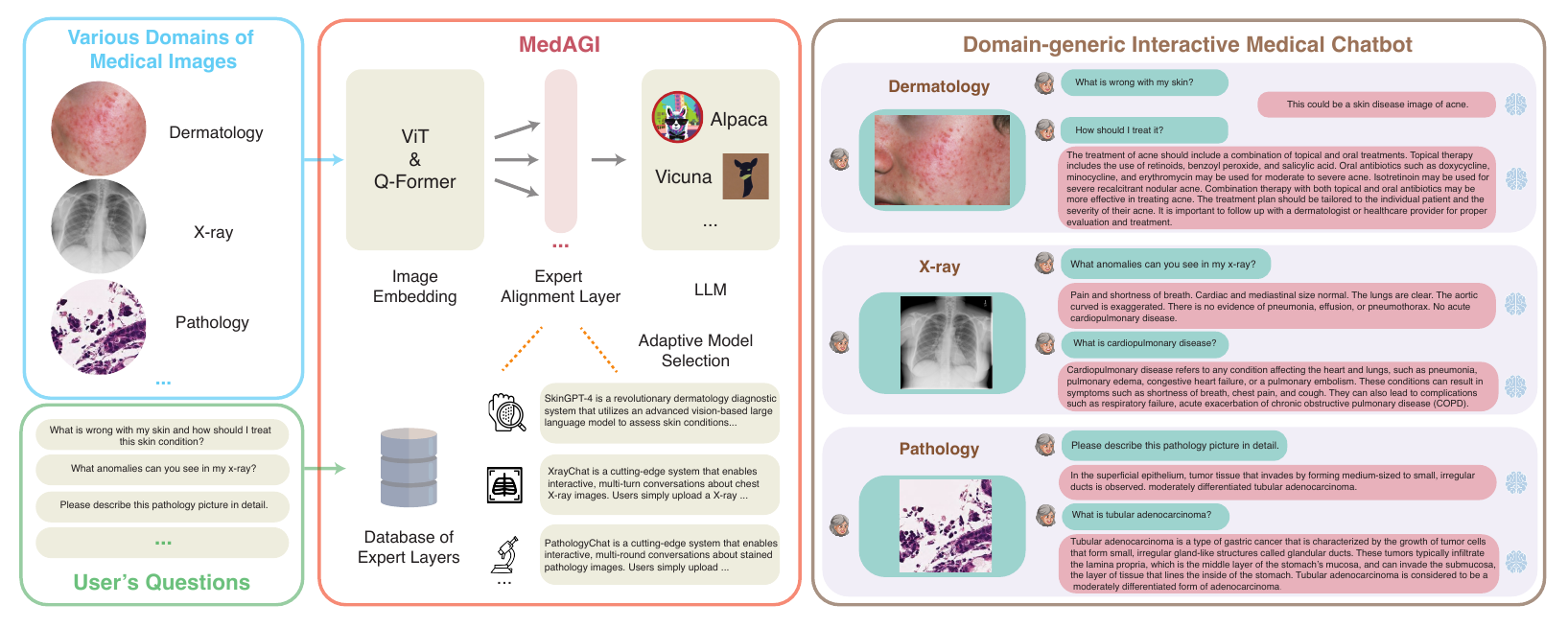}
    \caption{\textbf{Illustration of MedAGI.} MedAGI is a paradigm to unify domain-specific medical LLMs with the lowest cost. Users could upload images from any domain, such as dermatology, X-ray and pathology, and ask questions regarding the image. Then, MedAGI could automatically select the most suitable expert layer from the database by analyzing users' questions to provide the best response in the interactive diagnosis.}
    \label{fig_fig1}
\end{figure*}

One limitation of ChatGPT is its exclusive reliance on text input, with no support for direct image input. 
This absence of multimodal capabilities narrows its applicability in medical diagnosis, a field that often depends significantly on image-based data.
\cite{wang2023chatcad} tries to solve this problem in ChatCAD by integrating multiple image-text networks to transform medical imaging data, including X-rays, CT scans, and MRIs to textual description.
These transformed descriptions can subsequently be used as input for ChatGPT. 
% The model can then generate medical reports, enable interactive explanations, and offer pertinent medical recommendations, all based on the analyzed imagery.
However, the separation of the image-to-text transformation process from the LLMs process not only underutilizes the full potential of LLMs but can also lead to compromised performance if the quality of the image-to-text model is lacking.
Furthermore, it is crucial to address the potential data privacy concerns associated with ChatGPT's API for uploading text descriptions, as both medical images and textual patient information are highly sensitive\cite{li2023multi, lund2023information, sallam2023chatgpt}. 
To ensure the protection of patient confidentiality, careful consideration should be given to implementing robust privacy protection measures \cite{rajpurkar2022ai, zhou2022ppml, zhou2023personalized, zhou2023audit}.  

To solve the above two challenges, a number of open-source multimodal LLMs were proposed\cite{li2023blip,gao2023llama,ye2023mplug,zhu2022minigpt4,zhang2023video, dai2023instructblip, li2023videochat, gong2023multimodal, alayrac2022flamingo, sung2022vl, driess2023palm, huang2023language, zhang2023multimodal, koh2023grounding, liu2023visual, koh2023generating, chen2023x, li2023otter, wang2023interactive,li2023lmeye, zhang2023video, su2023pandagpt, girdhar2023imagebind}.
% , such as BLIP-2\cite{li2023blip}, LLaMA-Adapter V2\cite{gao2023llama}, mPLUG-Owl\cite{ye2023mplug}, MiniGPT-4 \cite{zhu2022minigpt4}, Video-LLaMA\cite{zhang2023video}, InstructBLIP\cite{dai2023instructblip}, VideoChat\cite{li2023videochat}, MultiModal-GPT\cite{gong2023multimodal}, and so on \cite{alayrac2022flamingo, sung2022vl, driess2023palm, huang2023language, zhang2023multimodal, koh2023grounding, liu2023visual, koh2023generating, chen2023x, li2023otter, li2023lmeye, zhang2023video, su2023pandagpt, girdhar2023imagebind}.  
In the medical field, there are two main approaches being adopted. 
The first involves training an end-to-end large multimodal model that combines a vision encoder and an LLM for visual and language understanding, such as LLaVA-Med\cite{li2023llava} and PathAsst\cite{sun2023pathasst}.
This strategy often faces challenges due to the need to gather data from various domains, which is especially challenging in medicine due to privacy issues and the lack of open-source datasets. 
The second approach seeks to bridge the gap between LLMs and pre-trained image encoders using an additional alignment layer, which is then fine-tuned using domain-specific data. 
This method, as employed in models such as SkinGPT-4\cite{zhou2023skingpt}, ProteinChat\cite{guo2023proteinchat}, XrayGPT\cite{thawkar2023xraygpt} and XrayChat\cite{liang2023xraychat}, is more feasible due to the requirement of fewer instances to fine-tune fewer parameters.

It's optimistic to envision that, in the future, an increasing number of domain-specific professional multimodal LLMs in the medical field will be developed. 
However, having them dispersed across various platforms, each with their own instructions, and leaving it up to users to find the model that fits their specific needs could be quite costly. 
It is also costly in terms of storage and loading resources to repeatedly store and load the same image encoder and language models for different multimodal LLMs.
Merging these models to form a universal medical model by using all the collected data is also unrealistic, given that medical data is typically non-public and not shared.
As an alternative, integrating these models into a unified platform could prove to be a powerful solution.
Hence, in this work, we propose Medical AGI (MedAGI), a paradigm to unify domain-specific medical LLMs with the lowest cost, and suggest a possible path to achieve medical AGI. 
Specifically, our MedAGI system is designed to automatically select appropriate medical models by analyzing users' questions. This selection process leverages the detailed descriptions of different medical models provided in their respective introductions, ensuring the best fit for the user's requirements.
In addition to saving space and being user-friendly, our model also boasts extendability. 
It doesn't require retraining, regardless of the number of new models proposed, making it a future-proof solution in the rapidly evolving field of medical AI.
% With the second strategy aforementioned, a large number of domain-specific alignment layers have been proposed in the community for multimodal LLMs.
% More alignment layers are on their way to being trained, and only the trained alignment layers could be open-sourced but not the raw medical data due to privacy issues. 
% In such a context, collecting data across domains to train a unified multimodal LLM is difficult to achieve, while MedAGI provides a unified interface to unite these existing domain-specific alignment layers and could automatically select the most appropriate domain-specific alignment layer based on the user's instruction with our adaptive model selection algorithm, 
% achieving AGI-like capability for medical diagnosis purposes, where users no longer need to care which domain their input image belongs to. 
To demonstrate the robustness of MedAGI, we evaluated it in three medical domains, including dermatology diagnosis, X-ray diagnosis, and analysis of pathology pictures. 
Our experiments revealed that MedAGI is proficient at selecting the appropriate models to match various user requirements. 
% Furthermore, in a double-blind test, users demonstrated an overwhelming preference for our model.
In conclusion, MedAGI stands out as a versatile and easily scalable solution. 
As the community continues to develop and increase the number of alignment layers trained on various domains, MedAGI only needs to manage these alignment layers to deliver a domain-generic performance, making it a promising tool for the future.

\begin{figure*}[!htb]
    \centering
    \includegraphics[width=1\linewidth]{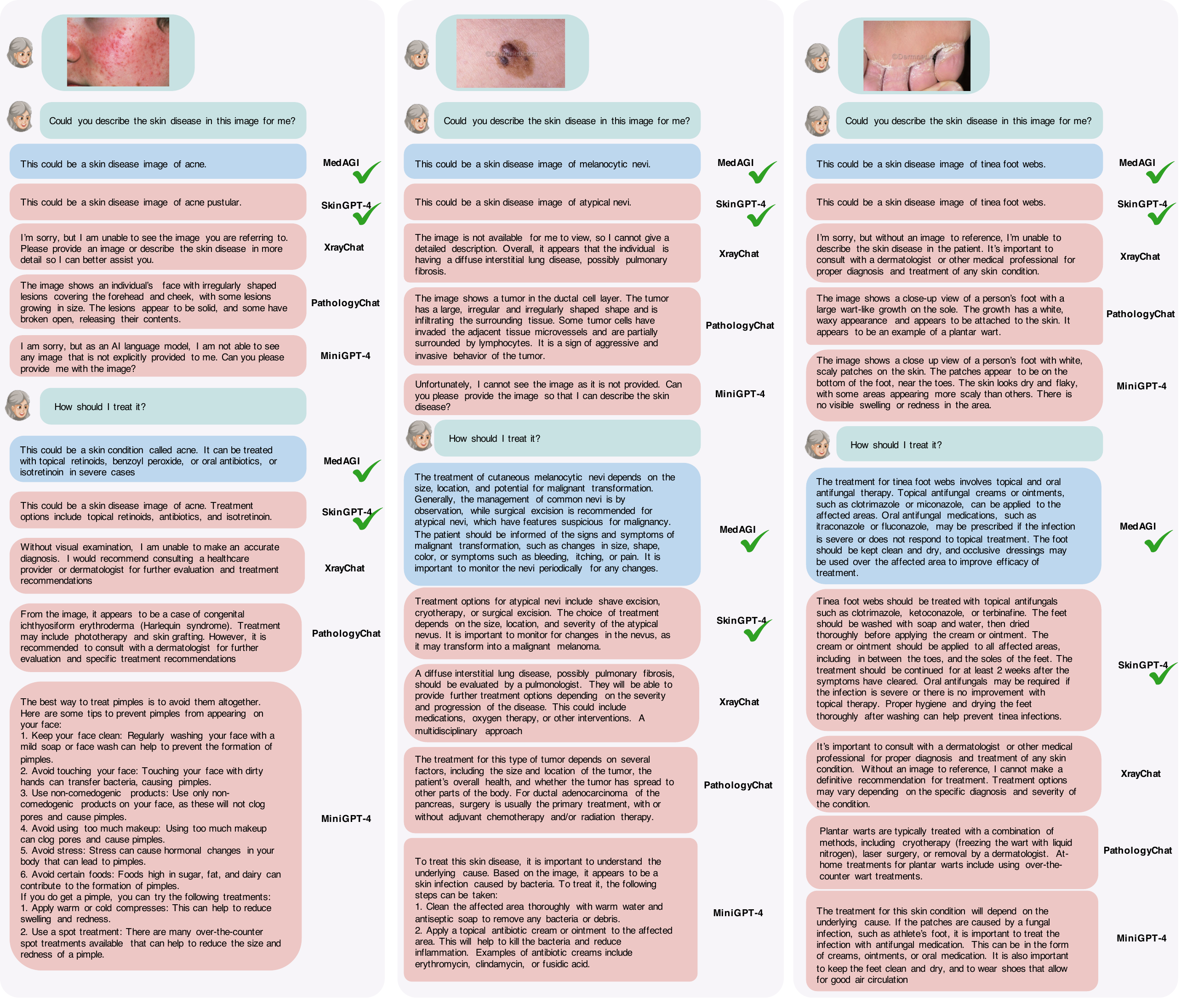}
    \caption{\textbf{Comparison of MedAGI, SkinGPT-4, XrayChat, PathologyChat, and MiniGPT-4 on three skin disease cases.} The green tick indicates that the answer is correct.}
    \label{fig_fig2}
\end{figure*}

\begin{figure*}[!htb]
    \centering
    \includegraphics[width=1\linewidth]{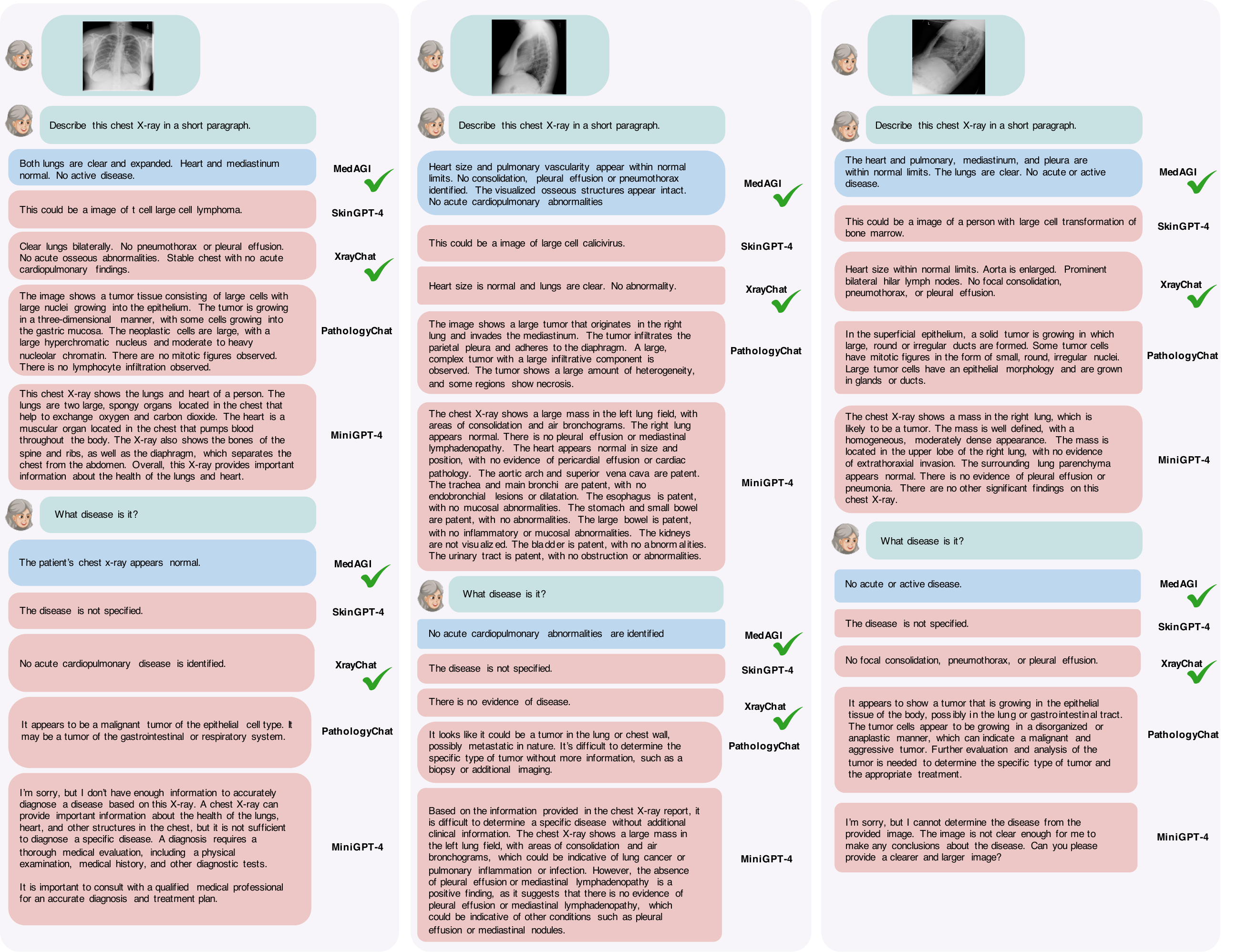}
    \caption{\textbf{Comparison of MedAGI, SkinGPT-4, XrayChat, PathologyChat, and MiniGPT-4 on three X-ray cases.} The green tick indicates that the answer is correct.}
    \label{fig_fig3}
\end{figure*}

\begin{figure*}[!htb]
    \centering
    \includegraphics[width=1\linewidth]{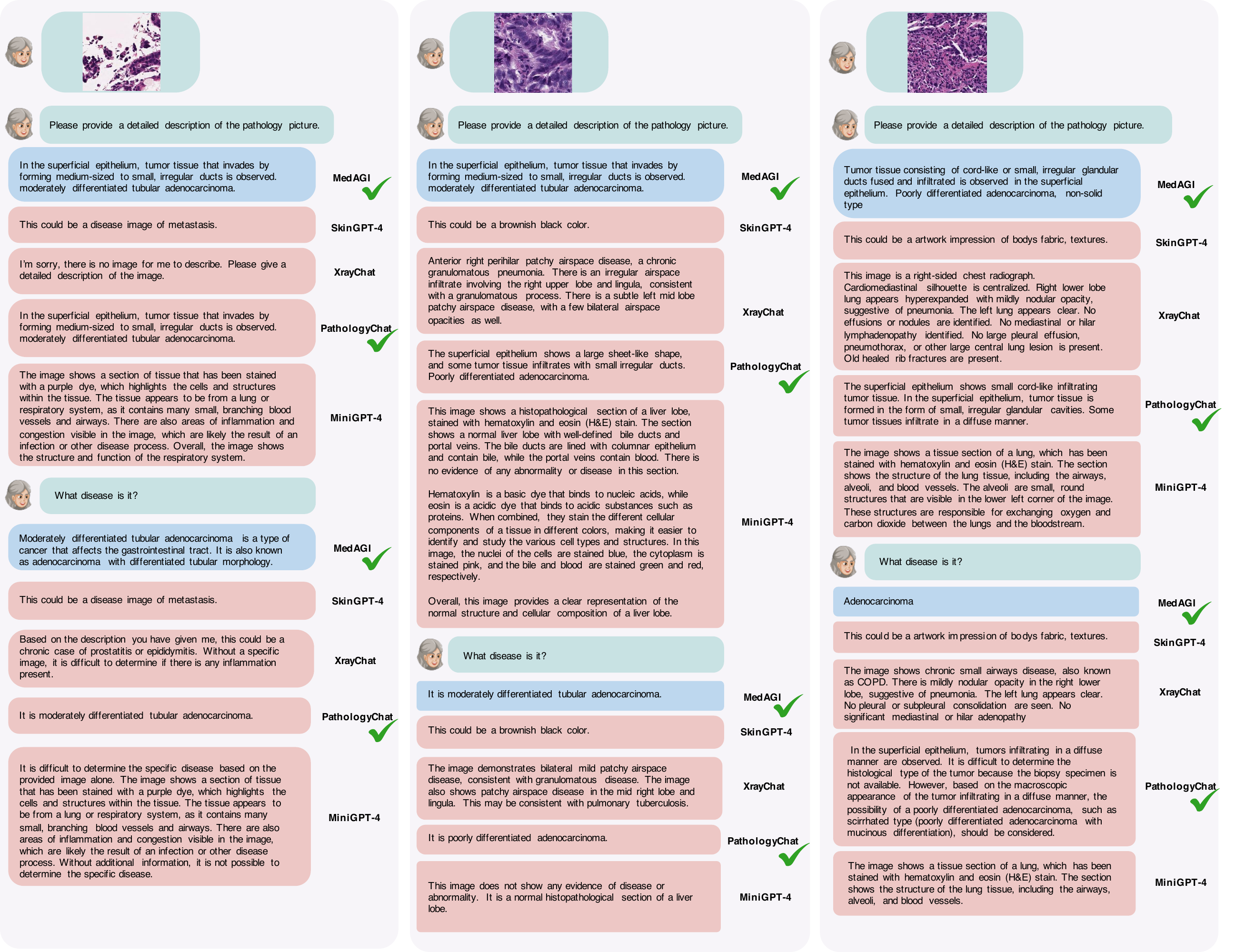}
    \caption{\textbf{Comparison of MedAGI, SkinGPT-4, XrayChat, PathologyChat, and MiniGPT-4 on three stained pathology cases.} The green tick indicates that the answer is correct.}
    \label{fig_fig4}
\end{figure*}

\section{Results}
\subsection{Design of MedAGI}
MedAGI is a paradigm to unify domain-specific medical LLMs with the lowest cost and a possible path to achieving medical AGI (\textbf{Figure} \ref{fig_fig1}).
By taking the user-uploaded image and user question as inputs, the system is capable of answering questions pertaining to different domains, including dermatology, X-ray analysis, and pathology, regarding the provided image.

Concretely, the uploaded image is first processed by the Vision Transformer (VIT)\cite{fang2022eva} and Q-Transformer models\cite{zhu2022minigpt4} for comprehensive understanding. 
The VIT model partitions the image into smaller patches and extracts crucial features. 
The Q-Transformer model then generates an embedding of the image by leveraging a transformer-based architecture, enabling the model to consider the image's contextual information.
Then MedAGI leverages the detailed descriptions of different medical models provided in their respective introductions stored in the database and selects the adaptive expert alignment layer in the domain-specific model that matches the user's intention the most.
The layer is then used to align the visual representation from Q-Transformer with the user question, enabling a coherent analysis of the image. 
Finally, the LLM utilizes the aligned information to generate a text-based diagnosis, providing a clear and concise description of the image corresponding to the user's question. 
Thus, MedAGI achieves AGI-like capability for medical diagnosis purposes, where users no longer need to care about which domain their input image belongs to.

\subsection{MedAGI Automatically Selects the Most Suitable Expert Layer}
In the absence of MedAGI, users are required to manually select the appropriate multimodal LLMs based on the specific image type and the manner in which they pose their questions. For example, they might have to choose between SkinGPT-4 for dermatology diagnosis, XrayChat for X-ray analysis, or PathologyChat for pathology image analysis, which adds complexity and necessitates domain-specific considerations.

Herein, MedAGI offers a unified interface that eliminates the need for users to worry about the specific domain to which a particular task belongs. It provides a seamless experience by integrating various domain-specific models into a single framework. To illustrate this, we conducted a comparative study involving MedAGI, SkinGPT-4, XrayChat, PathologyChat, and MiniGPT-4 across three domains, with three cases per domain, as depicted in Figure \ref{fig_fig2}-\ref{fig_fig4}.

As expected, SkinGPT-4, XrayChat, and PathologyChat performed well within their respective domains. SkinGPT-4 excelled in dermatology diagnosis, XrayChat showed proficiency in X-ray analysis, and PathologyChat demonstrated effectiveness in pathology image analysis. However, when faced with cross-domain scenarios, these domain-specific models exhibited limitations due to their lack of cross-domain knowledge and expertise.

In contrast, MedAGI proved capable of providing accurate and appropriate answers to user queries, even in cross-domain situations. This highlights MedAGI's domain-agnostic nature and its ability to handle a wide range of medical tasks, transcending specific domains.

\subsection{Scalability of MedAGI}
The scalability of MedAGI extends beyond the domains of dermatology diagnosis, X-ray analysis, and pathology image analysis. MedAGI's design allows for easy extension to a wide range of medical domains, making it a scalable solution for various healthcare applications. For instance, MedAGI could be seamlessly applied to domains such as cardiology, neurology, radiology, oncology, and many others. By leveraging domain-specific medical LLMs and incorporating them into the MedAGI framework, the system could analyze and interpret data from diverse medical specialities. This scalability enables healthcare professionals to access MedAGI's domain-generic capabilities across a broader spectrum of medical disciplines.

\section{Methods}
\subsection{Data processing and model training}
To demonstrate the robustness of MedAGI, we evaluated it in three medical domains, including dermatology diagnosis, X-ray diagnosis, and analysis of pathology pictures by implementing SkinGPT-4, XrayChat, and PathologyChat with MiniGPT-4 as the backbone, and gathering the expert alignment layer from them.

To implement SkinGPT-4, we followed the procedures demonstrated in \cite{zhou2023skingpt} and used two public datasets (SKINCON\cite{daneshjou2022skincon} and Dermnet) and a private in-house dataset, where the public datasets were used for the step 1 training, and the second public dataset and our in-house dataset were used for the step 2 training.

To implement XrayChat, we followed the procedures demonstrated in \cite{liang2023xraychat} and used 400K chest X-ray images and instructions, from Open-i and MIMIC CXR\cite{johnson2019mimic}.

To implement PathologyChat, we collected 262,777 patches extracted from 991 H\&E-stained gastric slides with Adenocarcinoma subtypes paired with captions extracted from medical reports\cite{tsuneki2022inference}.

During the training of both steps, the max number of epochs was fixed to 5, the iteration of each epoch was set to 5000, the warmup step was set to 5000, batch size was set to 2, the learning rate was set to 1e-4, and max text length was set to 160. The entire fine-tuning process required approximately 9 hours to complete and utilized two NVIDIA V100 (32GB) GPUs. The training was conducted on a workstation equipped with 252 GB RAM, 112 CPU cores, and two NVIDIA V100 GPUs.

\subsection{Algorithm for Adaptive Selection of Expert Alignment Layers}
Our expert alignment layer selection considers both the user question and different model instructions.
The model description is derived from the abstract of the corresponding paper. 
Formally, we represent the user input as $q=\{w^q_{1}, \cdots, w^q_{L_q}\}$, where $w^q_i$ is the $i$-th word, and $L_q$ is the input length.
Similarly, the $j$-th model description is denoted as $d=\{w^{d,j}_1,\cdots,w^{d,j}_{L_d}\}$.
We employ a BERT model pre-trained on 215M question-answering pairs from diverse sources\cite{reimers2019sentence} to encode each word sequence:
\begin{align}
    \begin{aligned}
\left\{\mathbf{h}^q_i, \cdots, \mathbf{h}^q_{L_q}\right\} & =\operatorname{Enc}\left(w^q_{1}, \cdots, w^q_{L_q}\right), \\
\left\{\mathbf{h}^{d,j}_i, \cdots, \mathbf{h}^{d,j}_{L_d}\right\} & =\operatorname{Enc}\left(w^{d,j}_{1}, \cdots, w^{d,j}_{L_d}\right).
\end{aligned}
\end{align}
where Enc is the encoder module in BERT which outputs the vector representation $\mathbf{h}^q_i$ of each input token $w^q_{1}$ in user input and $\mathbf{h}^{d,j}_i$ of each input token $w^{d,j}_{1}$ in the $j$-th model description.
To obtain a vector representation of the user input, we apply the mean-pooling operation to the hidden states of tokens:
\begin{align}
    u=\text { Mean-pooling }\left(\left\{\mathbf{h}^d_i, \cdots, \mathbf{h}^d_{L_d}\right\}\right).
\end{align}
The $j$-th model description is obtained as similarly $v^j$.
At inference, when predicting similarities between the two inputs, only the sentence embeddings $u$ and $v^j$ are used in combination with cosine-similarity:
\begin{align}
    s^j=\operatorname{similarity}(u,v^j). 
\end{align}
The model that obtains the highest $s$ score will be selected as the answering model.

The descriptions of SkinGPT-4, XrayChat, and PathologyChat in MedAGI were set as below:

\textbf{SkinGPT-4: }\textit{SkinGPT is a revolutionary dermatology diagnostic system that utilizes an advanced vision-based large language model to assess skin conditions. By uploading personal skin photos to the system, users receive an autonomous analysis that can identify and categorize various skin conditions, and provide treatment recommendations.}

\textbf{XrayChat: }\textit{XrayChat is a cutting-edge system that enables interactive, multi-turn conversations about chest X-ray images. Users simply upload a chest X-ray image, ask any question about it, and XrayChat generates informed responses. The system utilizes an X-ray encoder, a large language model, and an adaptor to comprehend the X-ray image and produce accurate and helpful answers.}

\textbf{PathologyChat: }\textit{PathologyChat is a cutting-edge system that enables interactive, multi-round conversations about stained pathology images. Users simply upload a pathology image, ask any question about it, and PathologyChat generates informed responses.}

\section{Conclusion and Discussion}
With the increasing number of domain-specific professional multimodal LLMs in the medical field, combining these models into a unified platform could prove to be a meaningful task. MedAGI is one of the possible solutions to unify domain-specific medical LLMs with the lowest cost.

In conclusion, MedAGI presents a promising paradigm for unifying domain-specific medical large language models (LLMs). By automatically selecting appropriate medical models based on users' questions, MedAGI eliminates the need for users to navigate multiple platforms and instructions, reducing costs and improving user experience. MedAGI represents a significant step towards the realization of medical artificial general intelligence. Its unified approach, scalability, and adaptability make it a compelling solution for the future of medical AI.

\section{Acknowledgements}
\noindent
\textbf{Funding: }Juexiao Zhou, Xiuying Chen, and Xin Gao were supported in part by grants from the Office of Research Administration (ORA) at King Abdullah University of Science and Technology (KAUST) under award number FCC/1/1976-44-01, FCC/1/1976-45-01, REI/1/5202-01-01, REI/1/5234-01-01, REI/1/4940-01-01, RGC/3/4816-01-01, and REI/1/0018-01-01.\\

\noindent
\textbf{Competing Interests: }The authors have declared no competing interests.\\

\noindent
\textbf{Author Contribution Statements: }J.Z., X.C. and X.G. conceived of the presented idea. J.Z. and X.C. designed the computational framework and analysed the data. X.G. supervised the findings of this work. J.Z., X.C. and X.G. took the lead in writing the manuscript. All authors discussed the results and contributed to the final manuscript.\\

\noindent
\textbf{Data availability: }The data for pathology can be accessed at \url{https://github.com/masatsuneki/histopathology-image-caption}. The data for XrayChat can be accessed at \url{https://github.com/UCSD-AI4H/xraychat}. The SKINCON dataset can be accessed at \url{https://skincon-dataset.github.io/}. The Dermnet dataset can be accessed at \url{https://www.kaggle.com/datasets/shubhamgoel27/dermnet}. The restricted in-house skin disease images of SkinGPT-4 are not publicly available due to restrictions in the data-sharing agreement.\\

\noindent
\textbf{Code availability: }The code proposed by MedAGI is publicly available at \url{https://github.com/JoshuaChou2018/MedAGI}.

{
\bibliographystyle{IEEEtran}
\bibliography{reg}
}

\end{document}